\pdfoutput=1

\documentclass[11pt]{article}

\usepackage[]{ACL2023}

\usepackage{times}
\usepackage{latexsym}

\usepackage[T1]{fontenc}

\usepackage[utf8]{inputenc}

\usepackage{microtype}

\usepackage{inconsolata}

\usepackage{custom} 

\title{KoBigBird-large: Transformation of Transformer\\for Korean Language Understanding}

\makeatletter
\renewcommand*{\@fnsymbol}[1]{\ensuremath{\ifcase#1\or \dagger\or \ddagger\or \mathsection\or \mathparagraph\or \|\or **\or \dagger\dagger\or \ddagger\ddagger \else\@ctrerr\fi}}
\makeatother
\author{
    Kisu Yang$^{a,b}$,
    Yoonna Jang$^{b}$,
    Taewoo Lee$^{a}$,
    Jinwoo Seong$^{a}$, \\
    {\bf Hyungjin Lee$^{a}$},
    {\bf Hwanseok Jang$^{a}$},
    {\bf Heuiseok Lim$^{b}$\thanks{~ Corresponding author} } \\
    $^{a}$VAIV Company AI Lab \\
    $^{b}$Korea University NLP\&AI Lab \\
    \texttt{\{ksyang, twlee, jinw.seong, hjlee, hsjang\}@vaiv.kr} \\
    \texttt{\{willow4, morelychee, limhseok\}@korea.ac.kr}}

\begin{document}
\maketitle

\begin{abstract}
This work presents KoBigBird-large, a \textit{large} size of Korean BigBird that achieves state-of-the-art performance and allows long sequence processing for Korean language understanding.
Without further pretraining, we only transform the architecture and extend the positional encoding with our proposed Tapered Absolute Positional Encoding Representations (TAPER).
In experiments, KoBigBird-large shows state-of-the-art overall performance on Korean language understanding benchmarks and the best performance on document classification and question answering tasks for longer sequences against the competitive baseline models.
We publicly release our model here\footnote{\url{https://huggingface.co/vaiv/kobigbird-roberta-large}}.
\end{abstract}

\section{Introduction}
Research on minority languages is a crucial area that extends beyond the scope of English language representations. Even though multilingual models for natural language understanding (NLU) \cite{devlin2019bert, conneau2020unsupervised} have widely shown moderate performance, they still fall short of expectations for Korean NLU tasks. This deficiency highlights the need to develop language-specific models that can effectively handle the distinctive characteristics of the language.

Currently, several Korean NLU models \cite{lee2020kr, park2020koelectra} have been proposed, among which KLUE-RoBERTa \cite{park2klue} has shown promising performance on general tasks.
It has been designed to reflect the characteristics of the language and pretrained with various model sizes including a \textit{large} size.
Despite its strengths, the vanilla Transformer-based architecture prevents itself from processing long sequences over a certain length, which affects its overall performance because important information could be lost \cite{beltagy2020longformer}.

Following BigBird which is proposed to process longer inputs \cite{zaheer2020big}, a Korean version of BigBird has been publicly released with a \textit{base} size \cite{jangwon_park_2021_5654154}.
However, owing to the limited size, it fails to match the performance of other \textit{large} models.
As a result, the absence of a \textit{large} size of Korean BigBird forces open-source users to choose either the competitive performance or long text processing although both are desirable features \cite{lee2022littlebird}.

\begin{figure}
  \includegraphics[width=\linewidth]{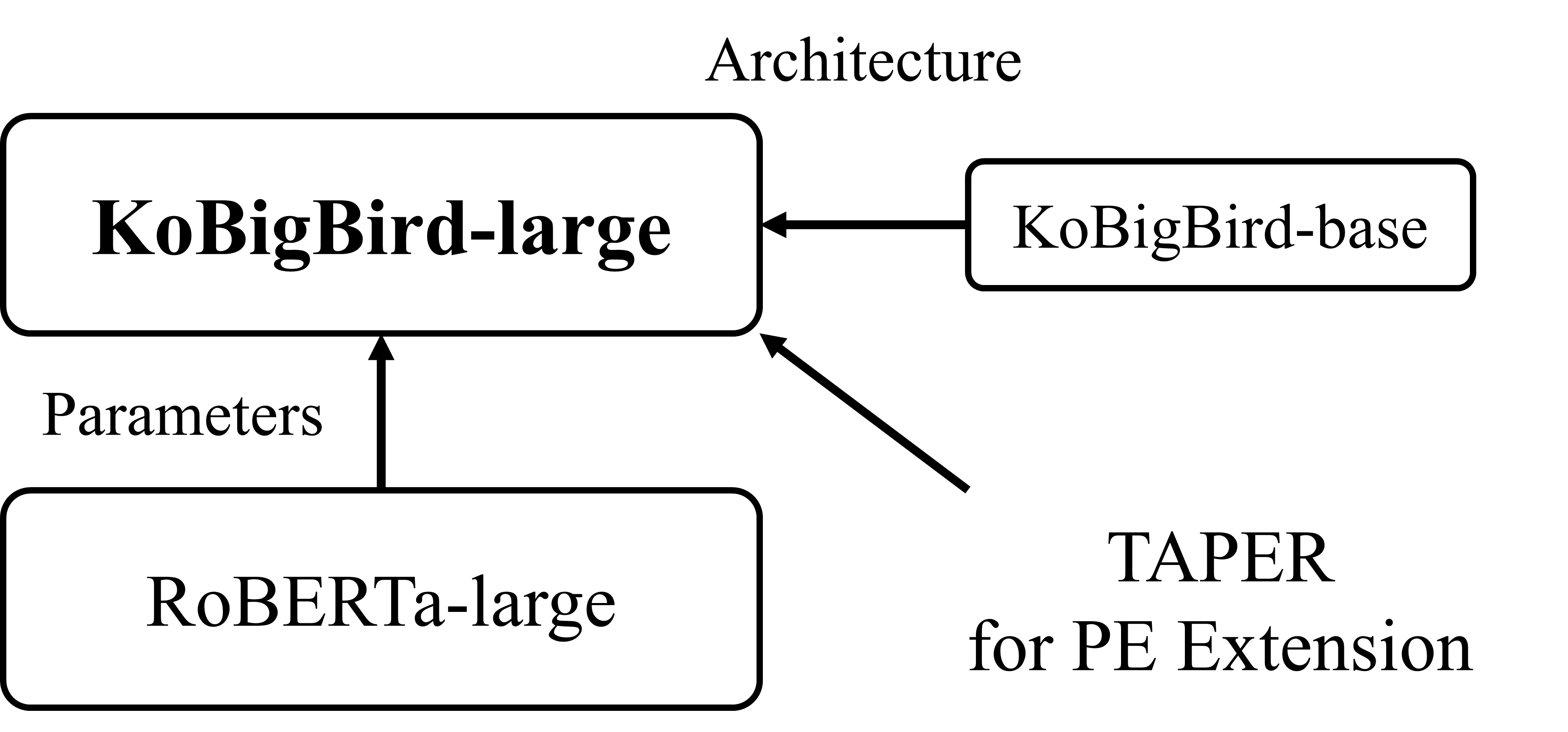}
  \caption{An illustration of building KoBigBird-large process. Based on the architecture of KoBigBird-base and the parameters of RoBERTa-large, our proposed TAPER method is applied to build KoBigBird-large.}
  \label{fig:concept}
\end{figure}

To overcome this limitation, we present KoBigBird-large, a \textit{large} size of Korean BigBird to simultaneously achieve state-of-the-art performance and longer sequence processing for Korean NLU tasks.
It is initialized with the \textit{large} size of KLUE-RoBERTa to
take advantage of the mentioned strengths and then transformed into the BigBird architecture with the Tapered Absolute Positional Encoding Representations (TAPER) which could extend position embeddings.

Noteworthily, no further pretraining or corpus is required to build it, but just the transformation of modules is all we need.
Details regarding the modifications for short sequences are described in Section \ref{subsec:full_attention_mode} while the methodology to improve extrapolation for extended sequences is in Section \ref{subsec:sparse_attention_mode}. 
This approach without further pretraining allows us to clarify the impact of differences in model structure and reduce carbon footprints in line with current ethical issues \cite{patterson2021carbon}.

In experiments, KoBigBird-large achieves state-of-the-art overall performance for all tasks on Korean NLU benchmarks \cite{park2klue} with an average gap of more than 0.4\% points compared to the previous records.
Also, the perplexity measurement of KoBigBird-large for longer inputs demonstrates that TAPER helps to improve the extrapolation of language models \cite{press2021train}. In additional experiments for long sequences, ours performs best on document classification \cite{nikl2020moducorpus} and question answering tasks against the competitive baseline models.
More details for experimental results are described in Section \ref{sec:experiments} and \ref{sec:analysis}.


\section{Related Work}
\paragraph{KLUE-RoBERTa}
Inspired by RoBERTa \cite{liu2019roberta}, KLUE-RoBERTa \cite{park2klue} has been proposed for Korean language processing.
As a pretraining corpus, a subset of ten corpora has been selected based on criteria such as diversity, modernity, privacy, or toxicity concerns.
The corpus has been pseudonymized with the Faker\footnote{\url{https://github.com/joke2k/faker}} library and morphologically analyzed \cite{kudo2005mecab} before its tokenizer has constructed the vocabulary using byte-pair encoding \cite{sennrich2015neural}.
The \textit{large} size of KLUE-RoBERTa has been pretrained with a batch size of 2048 and a fixed learning rate of 1e-4 for 500k steps.

In short, its strength lies in the use of a qualified corpus and a tokenizer with expensive pertaining, rather than the architecture itself.
Thus, we transplant its well-tuned parameters into ours and promote the architecture to be KoBigBird-large.

\paragraph{KoBigBird-base}
BigBird \cite{zaheer2020big} is a pretrained model that has been proposed to handle longer sequences. With dilated sliding window attention and different window sizes across the layers, BigBird allows the model to process 8 times longer tokens than BERT \cite{devlin2019bert}. It improves the computational efficiency in the long text by replacing the self-attention layer with sparse attention.

Based on English BigBird, KoBigBird-base, a Korean pretrained model for longer sequences, has been released \cite{jangwon_park_2021_5654154}. It is pretrained on multiple corpora encompassing Korean public corpus\footnote{\url{https://corpus.korean.go.kr/}}, Korean Wikipedia\footnote{\url{https://dumps.wikimedia.org/kowiki/}} and Common Crawl\footnote{\url{https://commoncrawl.org/}}. They adopt the WordPiece tokenizer and start pretraining from their own pretrained BERT weights. The model employs a batch size of 32 and a max sequence length of 4096, alongside a peak learning rate of 1e-4 which is coupled with a warmup phase of 20k steps, amounting to a total of 2M steps. The optimization is handled by the AdamW optimizer \cite{loshchilov2017decoupled}. The total number of pretraining tokens is less than that of KLUE-RoBERTa. Moreover, it is only available in the \textit{base} size and does not incorporate morpheme analysis, which hinders its practical use.

\section{Transformation of Transformer}
In this section, we provide a detailed description of our KoBigBird-large as a target model $M_{tgt}$ transformed from a source model $M_{src}$.
We choose KLUE-RoBERTa-large as $M_{src}$ because it features a morpheme-aware tokenizer tailored to Korean language characteristics, has been pretrained on ethically curated corpora, and stands out for its performance among language models for general Korean NLU tasks.

When the input length is the same as or shorter than the predefined length $l_{src}$ of $M_{src}$, KoBigBird-large operates in the full attention mode. During this mode, details of the embeddings and structure are upgraded to foster improvements in performance after fine-tuning, while ensuring output consistency with $M_{src}$ at the initial state. Details in this mode are elaborated in Section \ref{subsec:full_attention_mode}.

On the other hand, $M_{tgt}$ operates in the sparse attention mode when the input length exceeds $l_{src}$. In this case, since $M_{tgt}$ has an expanded input length $l_{tgt}$ that is greater than $l_{src}$, this extension defaults to generate randomly initialized $l_{tgt} - l_{src}$ absolute position embeddings (APE) unless otherwise handled. If used as is, they would provide inappropriate representations for positions.

To mitigate this problem, we propose a novel method, Tapered Absolute Positional Encoding Representations (TAPER), for the extended APEs so that they show better extrapolation performance for language modeling. The sparse attention mode with our newly proposed method is in Section \ref{subsec:sparse_attention_mode}. Model hyperparameters are provided in Table \ref{tab:hparam_arch}.

\subsection{Full Attention Mode}
\label{subsec:full_attention_mode}

KoBigBird-large incorporates an enhanced version of the embeddings and structure, initially adopting all parameters from $M_{src}$. Despite the modifications, it has been structured to ensure that, at the initial state, identical inputs yield consistent logits when compared to those produced by $M_{src}$.
Differences detailed below and further specifics can be verified in the released implementation.

\paragraph{Distinct Segment Type}
$M_{src}$ restricts the number of segment types to one as it is only trained with Masked Language Modeling (MLM) without Next Sentence Prediction (NSP). However, distinguishing separate input sentences can be significant in tasks such as Semantic Textual Similarity (STS), Natural Language Inference (NLI), and Machine Reading Comprehension (MRC). Thus, we created two segment type embeddings, duplicating the first segment type embeddings for the second. As the segment type embeddings remain untrained even during the pretraining of $M_{src}$, they are constituted by zeros.

\paragraph{Revised Positional Encoding}
When constructing absolute position embeddings for KoBigBird-large, we adopted the $l_{src}$ APEs of $M_{src}$. However, a bug in the implementation of it is observed, which starts position id counting from $2$, neglecting position ids $0$ and $1$. Thus, during transformation, we extracted the $l_{src}$ APEs from the range $[2, l_{src}+2)$.

\paragraph{Ordered Layer Normalization}
While $M_{src}$ applies layer normalization before dropout in the word embedding layer, it adopts the dropout before layer normalization order elsewhere. To ensure a consistent architecture, we apply dropout before layer normalization to the word embeddings layer. This alteration results in different logit values from $M_{src}$ and $M_{tgt}$ for the same input during the training mode, leading to different learning processes. However, they still return the exactly same logit values for the same input during inference at the initial state because the dropout is off.

\subsection{Sparse Attention Mode}
\label{subsec:sparse_attention_mode}

For input lengths exceeding $l_{src}$, KoBigBird-large switches to the Internal Transformer Construction (ITC) mode \cite{zaheer2020big}. It incorporates global, sliding, and random attention to handle inputs up to $l_{tgt}$. Alongside such structural changes for extended inputs, the extension generates untrained additional $l_{tgt} - l_{src}$ APEs. To address this, we introduce the TAPER method, which extends the originally trained $l_{src}$ APEs by applying attenuation to generate additional $l_{tgt} - l_{src}$ APEs.

\paragraph{TAPER}

\begin{figure}
  \includegraphics[width=\linewidth]{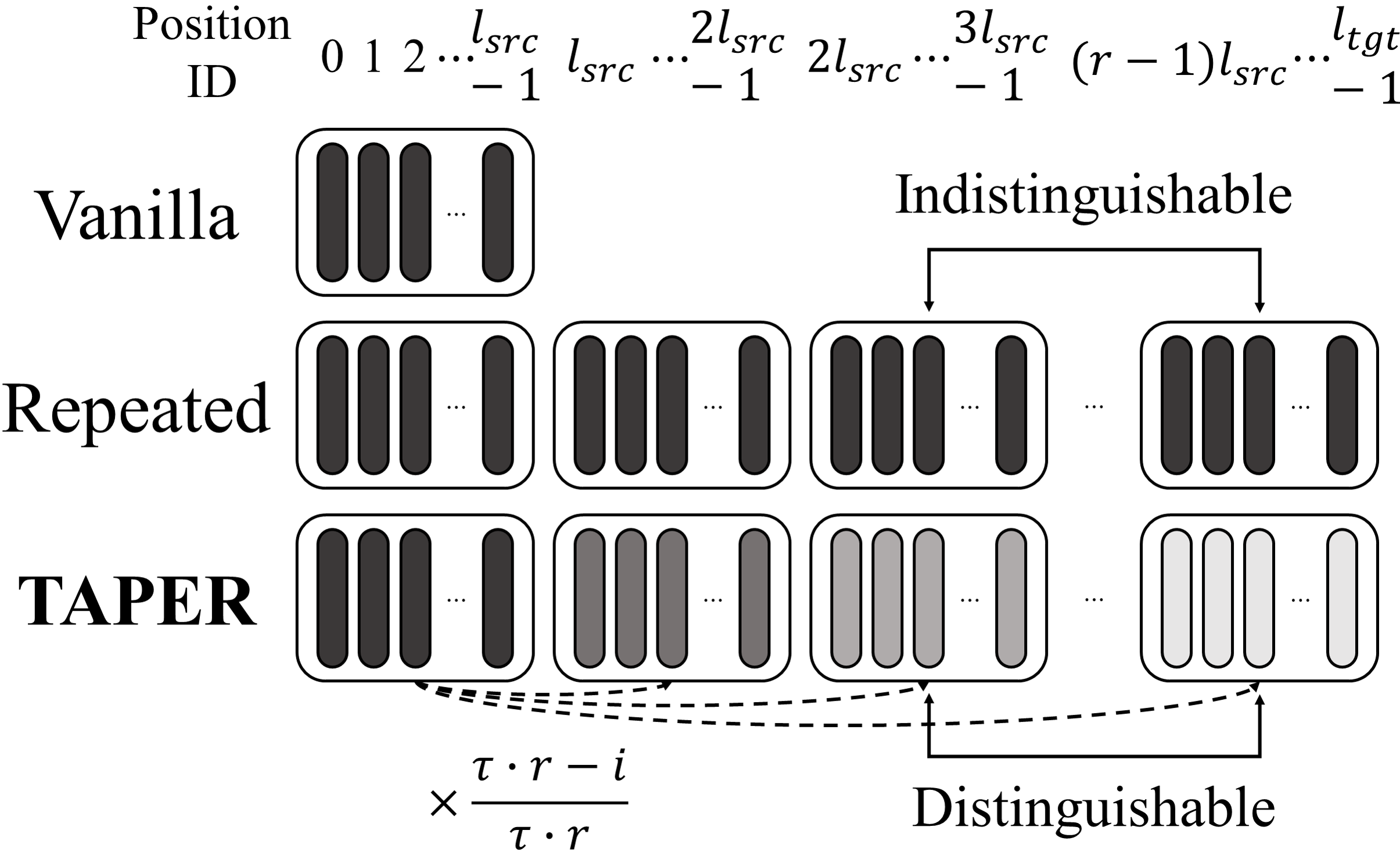}
  \caption{Illustrations showcasing the TAPER method's extension of position embeddings by multiplying the source embeddings with variables unique to each corresponding iteration, enhancing the informativeness and distinguishability of the extended positional representations.}
  \label{fig:taper}
\end{figure}

The motivation for TAPER stems from the operational characteristics of ALiBi \cite{press2021train}. When attending to a target token in an attention layer, ALiBi predominantly focuses on source tokens close to the target token. The attention score diminishes significantly with increasing distance, thereby having minimal influence on distant tokens.

Based on the characteristics, we hypothesize that even if the relationships with far-off tokens are unknown, repeating the pattern of pretrained position embeddings that are well-attuned to the relationships with nearby tokens should work well for a language modeling task.

However, simply repeating the pretrained $l_{src}$ APEs into an identical pattern $r = l_{tgt} / l_{src}$ times encounters the duplication problem: The APE for any arbitrary position $x$ becomes indistinguishable from the APE at position $x+l_{src}$. To address this, we apply the attenuated amplitude of each repetition.

\begin{equation}
\label{eq:taper}
    P_{tgt} = \displaystyle{\concat_{i=0}^{r-1} P_{src} \cdot \frac{\tau \cdot r - i}{\tau \cdot r} }
\end{equation}

In Equation \ref{eq:taper}, the source position embeddings $P_{src}$ are extended to the target position embeddings $P_{tgt}$.
The number of repetitions, $r$, represents an integer quotient of $l_{tgt}$ divided by $l_{src}$.
A temperature, $\tau$, determines the degree of attenuation applied to the amplitude of the APEs of each repetition, thereby making the extended positions distinguishable (see Figure \ref{fig:taper}). As $\tau$ increases, the difference in the APEs of each repetition diminishes, and if it becomes too large, the extended APEs become almost identical to the repeated ones.

We set $\tau=2.0$ for KoBigBird-large. The value of $\tau$ should be adjusted so that the extended position embeddings show the best extrapolation performance. The attenuated embeddings are concatenated along the dimension representing the position id.

\section{Experiments}
\label{sec:experiments}
\subsection{KLUE Benchmark}
\label{subsec:klue_benchmark}

The Korean Language Understanding Evaluation (KLUE) benchmark \cite{park2klue}, designed to foster Korean language processing research, encompasses eight varied Korean NLU tasks including topic classification, semantic textual similarity, and more. To enable equitable model comparisons in Korean NLP, KLUE provides benchmark datasets, task-specific evaluation metrics and pretrained language models like KLUE-RoBERTa. Data statistics are served in Table \ref{tab:klue_statistics}.

\paragraph{YNAT} This dataset comprises news headlines from online articles circulated by Yonhap News Agency, categorized into seven topics for Topic Classification (TC): politics, economy, society, culture, world, IT/science, and sports. Macro F1 score was adopted as the evaluation metric.

\paragraph{KLUE-STS} This dataset contains annotations of Semantic Textual Similarity (STS) between two sentences, rated from 0 to 5. Evaluation can be performed using the Pearson correlation coefficient between labels and predictions and the F1 score determined after converting them to binary using a threshold of 3.0.

\paragraph{KLUE-NLI} This Natural Language Inference (NLI) dataset includes pairs of sentences and corresponding labels of three types: entailment, contradiction, and neutral. These denote the relation between premise and hypothesis sentences. The performance is gauged by its classification accuracy.

\paragraph{KLUE-NER} This Named Entity Recognition (NER) dataset annotates entity classes for each character in a sentence, including person, location, organization, date, time, and quantity. The evaluation involves entity-level and character-level macro F1 scores.

\begin{table}[t]
\centering
\begin{adjustbox}{width=1\columnwidth}
\begin{tabular}{lrrrrr}
\toprule
\textBF{Dataset}  & \multicolumn{1}{c}{\textBF{|Train|}} & \multicolumn{1}{c}{\textBF{|Dev|}} & \multicolumn{1}{c}{\textBF{$L_{min}$}} & \multicolumn{1}{c}{\textBF{$L_{avg}$}} & \multicolumn{1}{c}{\textBF{$L_{max}$}} \\ \midrule
\textBF{YNAT}     & 45,678                      & 9,107                     & 4                          & 27                         & 44                         \\
\textBF{KLUE-STS} & 11,668                      & 519                       & 17                         & 66                         & 207                        \\
\textBF{KLUE-NLI} & 24,998                      & 3,000                     & 29                         & 70                         & 170                        \\
\textBF{KLUE-NER} & 21,008                      & 5,000                     & 25                         & 71                         & 222                        \\
\textBF{KLUE-RE}  & 32,740                      & 7,765                     & 17                         & 93                         & 432                        \\
\textBF{KLUE-DP}  & 10,000                      & 2,000                     & 16                         & 48                         & 140                        \\
\textBF{KLUE-MRC} & 26,128                      & 8,643                     & 209                        & 1,052                      & 2,070                      \\
\textBF{WoS}      & 8,000                       & 1,000                     & 110                        & 522                        & 1,429                      \\ \bottomrule
\end{tabular}
\end{adjustbox}
\caption{Data statistics of KLUE benchmark. It shows the number of samples and the minimum, average, and maximum length of input characters of each development set. We count the sum of two sentences for the tasks involving multiple sentences (KLUE-STS, KLUE-NLI and KLUE-MRC).}
\label{tab:klue_statistics}
\end{table}

\paragraph{KLUE-RE} This Relation Extraction (RE) dataset contains annotations for semantic relationships between subject and object entities in sentences. It includes 30 relationship classes, including a "no\_relation" label. Evaluation involves Micro F1 score without the "no\_relation" class and the Area Under the Precision-Recall Curve (AUPRC).

\paragraph{KLUE-DP}
This dataset annotates Dependency Parsing (DP) in sentences using syntax and function tags. The performance is evaluated using Unlabeled Attachment Score (UAS) for function tags and Labeled Attachment Score (LAS) for both function and syntax tags.

\paragraph{KLUE-MRC}
This Machine Reading Comprehension (MRC) dataset contains text, questions, and answer spans indicating where the answers are located in the text. The model's performance is evaluated based on the Exact Match (EM) of finding the answer location accurately at the character level and the ROUGE-W score, which measures the similarity between the predicted and actual answers using the longest common consecutive subsequence.

\begin{table*}[t]
\centering
\begin{adjustbox}{width=0.75\textwidth}
\begin{tabular}{lrrr}

\toprule

\multicolumn{1}{c}{\textBF{Parameter}} & \multicolumn{1}{c}{\textBF{KoBigBird-large}} & \multicolumn{1}{c}{KoBigBird-base} & \multicolumn{1}{c}{RoBERTa-large} \\ 

\midrule

Max. position embeddings      & 4096                               & 4096                              & 512                               \\
\# of segment embeddings      & 2                                   & 2                                  & 1                                 \\
Vocabulary size               & 32000                              & 32500                             & 32000                            \\
Hidden size                   & 1024                               & 768                                & 1024                             \\
Intermediate size             & 4096                               & 3072                              & 4096                             \\
Bias                          & True                                & True                               & True                              \\
Activation layer              & GELU                                & GELU$_{new}$                          & GELU                              \\
\# of heads                   & 16                                  & 12                                 & 16                                \\
\# of hidden layers           & 24                                  & 12                                 & 24                                \\
Dropout probability                  & 0.1                                 & 0.1                                & 0.1                               \\
Block length                  & 64                                  & 64                                 & -                                 \\
\# of random blocks           & 3                                   & 3                                  & -                                 \\

\bottomrule

\end{tabular}
\end{adjustbox}
\caption{Comparison of hyperparameters for representative Korean NLU models.}
\label{tab:hparam_arch}
\end{table*}

\paragraph{WoS}
Wizard of Seoul (WoS) is a Dialog State Tracking (DST) dataset that labels slot-value pairs in dialogues between humans (travelers) and computers (information sources). Slots represent categories (e.g., hotel type), while values represent possible options (e.g., hotel, guest house). The evaluation measures include Joint Goal Accuracy (JGA), which assesses if all slots are predicted accurately, and Slot F1 score, which measures the prediction accuracy for each individual slot.

\subsection{Baselines}

Apart from KLUE-RoBERTa and KoBigBird models, we additionally assess two multilingual language models and two Korean monolingual language models for the benchmark.

\paragraph{mBERT} This is a multilingual BERT model put forth and made publicly available. It is trained on a multilingual corpus that includes 104 languages, Korean included, using both the Masked Language Modeling (MLM) and Next Sentence Prediction (NSP) objectives \cite{devlin2019bert}.

\paragraph{XLM-R} A variant of RoBERTa trained on a vast multilingual corpus using the MLM objective \cite{liu2019roberta}.
 
\paragraph{KR-BERT} A publicly available Korean language model at the character level, based on BERT. The KR-BERT character WordPiece tokenizer incorporates a vocabulary of 16,424 unique tokens \cite{lee2020kr}.

\paragraph{KoELECTRA} An open-source Korean language model, trained with the MLM and replaced token detection objectives \cite{park2020koelectra}.

\subsection{Settings}

The learning rate of 2e-5 with the AdamW optimizer is used for all KLUE benchmark tasks, influenced by prior research in which adjustments were made in the range of 1e-5 to 5e-5, adopting a consistent value for easy reimplementation.
Instead, the batch size is selected from \{8, 16, 32\}.
Maximum sequence lengths are 128 for YNAT, KLUE-STS, KLUE-NLI, 256 for KLUE-RE and KLUE-DP, and 512 for KLUE-NER, KLUE-MRC, WoS. 

\subsection{Benchmark Results}

\begin{table*}[t]
\centering
\renewcommand{\arraystretch}{1.2}
\begin{adjustbox}{width=1\textwidth}
\begin{tabular}{l c cc c cc cc cc cc cc c}

        \toprule

        & \textBF{YNAT}              & \multicolumn{2}{c}{\textBF{KLUE-STS}}           & \textBF{KLUE-NLI}             & \multicolumn{2}{c}{\textBF{KLUE-NER}}           & \multicolumn{2}{c}{\textBF{KLUE-RE}}            & \multicolumn{2}{c}{\textBF{KLUE-DP}}            & \multicolumn{2}{c}{\textBF{KLUE-MRC}}           & \multicolumn{2}{c}{\textBF{WoS}} & \textBF{Overall} \\ \cmidrule(lr){2-2} \cmidrule(lr){3-4} \cmidrule(lr){5-5} \cmidrule(lr){6-7} \cmidrule(lr){8-9} \cmidrule(lr){10-11} \cmidrule(lr){12-13} \cmidrule(lr){14-15} \cmidrule(lr){16-16}
        
        \textBF{Model} & F1 & R$^P$ & F1 & ACC & F1$^E$ & F1$^C$ & F1$^{mic}$ & AUC & UAS & LAS & EM & ROUGE & JGA & F1$^S$ & AVG$^{mac}$\\

        \midrule

        \textBF{mBERT-base*}             & 82.64          & 82.97          & 75.93          & 72.90          & 75.56          & 88.81          & 58.39          & 56.41          & 88.53          & 86.04          & 49.96          & 55.57          & 35.27          & 88.60          & 72.07          \\
        \textBF{XLM-RoBERTa-base*} & 84.52 & 88.88 & 81.20 & 78.23 & 80.48 & 92.14 & 57.62 & 57.05 & 93.12 & \underline{87.23} & 26.76 & 53.36 & 41.54 & 89.81 & 73.42 \\

        \textBF{KR-BERT-base*}           & 85.36          & 87.50          & 77.92          & 77.10          & 74.97          & 90.46          & 62.83          & 65.42          & 92.87          & 87.13          & 48.95          & 58.38          & 45.60          & 90.82          & 75.49          \\
        \textBF{KoELECTRA-base*}         & 85.99          & \underline{93.14}    & 85.89          & 86.87          & \underline{86.06}    & \underline{92.75}    & 62.67          & 57.46          & 90.93          & 87.07          & 59.54          & 65.64          & 39.83          & 88.91          & 78.48          \\

        \midrule
        
        \textBF{KLUE-RoBERTa-base}                & \underline{86.62}    & 92.10          & 86.92          & 86.27          & 84.68          & 91.63          & 66.67          & 66.56          & 93.49          & 86.89    & 67.80          & 73.73          & 47.43          & 91.47  & 80.95                    \\
\textBF{KLUE-RoBERTa-large}               & 86.60          & 92.26          & \underline{87.01}    & \textBF{90.23} & 84.86          & 91.67          & \underline{69.83}    & \underline{72.44}    & \underline{93.86}    & 87.15          & \underline{75.35}    & \underline{80.97}    & \underline{47.69}    & \underline{91.52}    & \underline{83.02}          \\
        
        \midrule
        
        \textBF{KoBigBird-base}                & \textBF{86.84} & 92.36    & 86.54          & 87.03          & \textBF{87.17} & \textBF{92.90} & 65.96          & 64.36          & 93.54          & 86.61          & 67.32          & 73.37          & 47.49          & 91.37          & 81.05          \\
\textBF{KoBigBird-large (ours)}        & 86.40          & \textBF{93.69} & \textBF{88.37} & \underline{89.57}    & 85.09        & 91.93      & \textBF{70.49} & \textBF{72.65} & \textBF{94.19} & \textBF{87.44} & \textBF{75.57} & \textBF{81.44} & \textBF{50.07} & \textBF{91.94} & \textBF{83.43}          \\
        
        \bottomrule
        
\end{tabular}
\end{adjustbox}
\caption{Comparative experiments of our model to other Korean models on KLUE benchmark. The scores in \textbf{bold} indicate the best score, and the \underline{underline} indicate the second best score.  
}
\label{tab:short_klue}
\end{table*}
\begin{table}[t]
\centering
\begin{adjustbox}{width=1\columnwidth}
\begin{tabular}{c cc c cc}

        \toprule
        
        & \multicolumn{2}{c}{\textBF{KLUE-STS}}           & \textBF{KLUE-NLI}       & \multicolumn{2}{c}{\textBF{KLUE-MRC}} \\ \cmidrule(lr){2-3} \cmidrule(lr){4-4} \cmidrule(lr){5-6}
                \textBF{Segment Type} & R$^P$ & F1 & ACC & EM & ROUGE \\ 
                
        \midrule
        
        \textBF{Uniform}                & 93.36          & 87.58          & 89.53          & 75.04          & 81.07 \\
        \textBF{Distinct}               & \textBF{93.69}          & \textBF{88.37}    & \textBF{89.57} & \textBF{75.57}    & \textBF{81.44} \\

        \bottomrule
        
\end{tabular}
\end{adjustbox}
\caption{The experimental results based on different segment type embeddings. In the task of KLUE-STS, KLUE-NLI, and KLUE-MRC, distinct types achieve higher scores than uniform types.}
\label{tab:short_seg}
\end{table}

Table \ref{tab:short_klue} presents interesting results regarding the performance of different models on Korean NLU tasks. Performances marked with an asterisk (*) were borrowed from previous studies \cite{park2klue}, providing a valuable benchmark for comparison.

Firstly, multilingual models demonstrate weak performance in Korean NLU tasks, affirming the importance of language-specific models for these tasks. Among the \textit{base} size models, KLUE-RoBERTa-base and KoBigBird-base show superior performance compared to other publicly available Korean NLU models.

When comparing the \textit{base} and \textit{large} size models, there is a performance increase when scaling up. Specifically, KLUE-RoBERTa-large achieves a 2.07\% points improvement over its base counterpart, and our KoBigBird-large shows a 2.38\% points increase over KoBigBird-base. When comparing \textit{large} models, KoBigBird-large performs 0.41\% points better than KLUE-RoBERTa-large, demonstrating its potential for more advanced NLU tasks.

An interesting anomaly is observed in the YNAT dataset where, in line with previous research \cite{park2klue}, \textit{base} models outperform \textit{large} models. This might be due to the nature of the YNAT task, which involves classifying topics based on titles. The data for this task may be too easy to classify, leading to rapid overfitting during training with larger models.

\subsection{Effects of Distinct Segment Type}

Table \ref{tab:short_seg} presents findings on the role of segment type embeddings in our KoBigBird-large model which includes two segment type embeddings to distinguish between different types of text segments. These embeddings are initially identical, but they are separately adapted during fine-tuning.

Our experiments were conducted on tasks that require multiple text inputs, specifically Semantic Textual Similarity (STS), Natural Language Inference (NLI), and Machine Reading Comprehension (MRC) tasks.

The results in Table \ref{tab:short_seg} show performance differences depending on whether or not segment types are distinguished. It was observed that distinguishing segment types tended to offer performance advantages in tasks involving multiple text inputs.

This evidence indicates the potential benefit of employing distinct segment type embeddings in tasks with multiple text inputs, emphasizing the adaptability and flexibility of the KoBigBird-large model in handling complex NLU tasks.

\section{Analysis}
\label{sec:analysis}
\subsection{Extrapolation}
\label{subsec:extrapolation}

\begin{figure}
  \includegraphics[width=\linewidth]{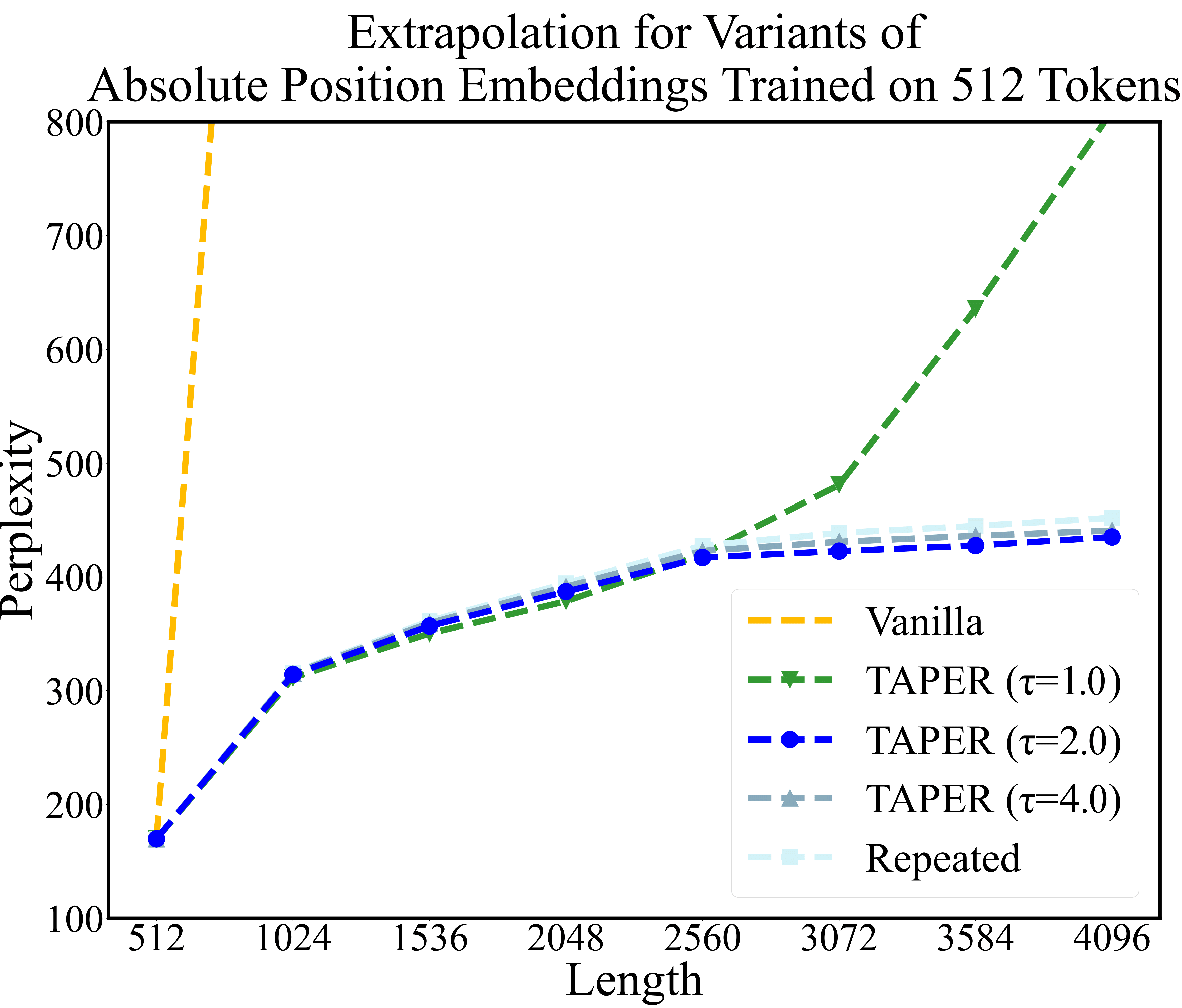}
  \caption{Perplexity scores on Korean Wikipedia corpus for extrapolation measurement}
  \label{fig:extrapolation}
\end{figure}

We investigate the impact of position embeddings and the temperature parameter $\tau$ on the model's perplexity (PPL) as in Figure \ref{fig:extrapolation}. As one of the metrics to evaluate language models, it represents the uncertainty of the model, so a lower score indicates higher performance.

We measure the PPL on preprocessed Korean Wikipedia corpus\footnote{\url{https://ratsgo.github.io/embedding/downloaddata.html}}. Inputs to a language model are packed with full sequences sampled contiguously from one or more documents, with separate tokens inserted between them to delimit individual documents like \citet{liu2019roberta}. Without random token replacement \cite{devlin2019bert}, 15\% of the input tokens are replaced with masked tokens. The PPL is measured only for these masked tokens, and the average is calculated.

In Figure \ref{fig:extrapolation}, "Vanilla" refers to the model prior to the application of TAPER, where the newly extended position embeddings are randomly initialized. On the other hand, "Repeated" signifies a scenario where the same value is repeated without the application of a temperature parameter ($\tau$).

Our findings show that the Vanilla model struggles to make predictions for lengths exceeding the pretraining limit, indicating the necessity of a more nuanced approach for longer sequences. When a temperature ($\tau$) of $1.0$ is applied, the PPL diverges after a certain point, suggesting a limit to the model's capacity to handle long sequences effectively in this configuration.

The model performs best with a temperature ($\tau$) of $2.0$, achieving the lowest PPL, suggesting that this temperature setting allows the model to handle longer sequences more effectively. Beyond this point, however, as the temperature increases, the PPL slightly rises, indicating that too high a temperature close to the repetition of APEs may have a negative impact on the model's performance.

These results suggest that careful tuning of the temperature parameter and adequate handling of position embeddings are crucial for optimizing the model's performance, particularly for long sequences.

\subsection{Long Text Classification}

In this part, we focus on longer text processing.
For a single text NLU task, we adopt the Sentiment Analysis dataset from Modu Corpus \cite{nikl2020moducorpus} built by the National Institute of the Korean Language. It consists of a total of 2,081 documents based on blogs or social media posts, and its topics are related to products, movies, and travel.
As represented in Table \ref{tab:modu_statistics}, we divide the total data into training and validation sets with a 4:1 ratio.
Although some documents exceed the maximum token length of the models, the models utilize tokens within a predetermined length.
Each document has five sentiment labels: strong negative, negative, neutral, positive, and strong positive.
Macro F1 score is used as the evaluation metric for the classification task.

Table \ref{tab:long_cls} presents the experimental results of the KLUE-RoBERTa and KoBigBird models for long document classification, including cases where the TAPER method is not applied to the KoBigBird-large as the ablation study. In our experiments, the KoBigBird-large not only achieves its peak performance at a sequence length of 512 but also retains the highest performance in longer sequences up to a length of 4096. This underscores the effective enhancements in both full attention mode and sparse attention mode. Scaling from the \textit{base} to the \textit{large} model size yields a notable performance increase of around 10\% points.

When we do not apply TAPER to the KoBigBird-large, and the extended position embeddings initiate randomly, the model suffers a significant performance decline even after fine-tuning. Strikingly, in sequences longer than 2048, the \textit{large} model without TAPER underperforms compared to the \textit{base} model, highlighting the substantial role of TAPER in sustaining performance. Overall, these results affirm the necessity of TAPER application in optimizing the performance of the KoBigBird-large, emphasizing its crucial role in handling longer sequences.

\begin{table}[t]
\centering
\begin{adjustbox}{width=1\columnwidth}
\begin{tabular}{crrrrrrrr}

\toprule

\textBF{Settype} & \multicolumn{1}{c}{\textBF{\#}} & \multicolumn{1}{c}{\textBF{Min}} & \multicolumn{1}{c}{\textBF{25\%}} & \multicolumn{1}{c}{\textBF{50\%}} & \multicolumn{1}{c}{\textBF{75\%}} & \multicolumn{1}{c}{\textBF{Max}} & \multicolumn{1}{c}{\textBF{Avg.}}    & \multicolumn{1}{c}{\textBF{Std.}}     \\

\midrule

\textBF{Train}   & 1,664                              & 19                      & 240                      & 312                      & 460                      & 10,383                   & 544 & 691  \\

\textBF{Dev}     & 417                               & 44                      & 248                      & 317                      & 446                      & 5,705                   & 500 & 597  \\

\bottomrule

\end{tabular}
\end{adjustbox}
\caption{Data statistics of Modu Sentiment dataset for long text classification. It presents the number of samples and the minimum, quartiles, and maximum lengths of input characters of each split.}
\label{tab:modu_statistics}
\end{table}

\begin{table}[t]
\centering
\begin{adjustbox}{width=1\columnwidth}
\begin{tabular}{lcccc}

        \toprule

        \multicolumn{5}{c}{\textBF{Single Document Classification}} \\

        \midrule

        \multicolumn{1}{c}{\textBF{Model}} & \textBF{512}                                & \textBF{1024}                            & \textBF{2048}                      & \textBF{4096}                      \\

        \midrule
        
        KLUE-RoBERTa-base         & 42.61                              & -                               & -                         & -                         \\
        KLUE-RoBERTa-large        & \underline{52.30}                              & -                               & -                         & -                         \\

        \midrule

        KoBigBird-base            & 45.81                              & 46.14                           & \underline{47.47}            & \underline{44.90}            \\

        KoBigBird-large (ours)    & \textBF{55.32}                     & \textBF{53.44}                  & \textBF{58.22}               & \textBF{52.06}               \\
        $-$ TAPER                   & - & \underline{47.44} & 43.58 & 43.33 \\

        \bottomrule        

\end{tabular}
\end{adjustbox}
\caption{Macro F1 scores for the 5-class classification task on Modu Sentiment datasets. Models perform sentiment analysis with various token lengths.}
\label{tab:long_cls}
\end{table}

\begin{table}[t]
\centering
\begin{adjustbox}{width=1\columnwidth}
\begin{tabular}{crrrrrrrr}

\toprule

\textBF{Settype} & \multicolumn{1}{c}{\textBF{\#}} & \multicolumn{1}{c}{\textBF{Min}} & \multicolumn{1}{c}{\textBF{25\%}} & \multicolumn{1}{c}{\textBF{50\%}} & \multicolumn{1}{c}{\textBF{75\%}} & \multicolumn{1}{c}{\textBF{Max}} & \multicolumn{1}{c}{\textBF{Avg.}}    & \multicolumn{1}{c}{\textBF{Std.}}     \\

\midrule

\textBF{Train}   & 17,554                              & 504                      & 727                      & 940                      & 1,299                      & 2,100                   & 1,037 & 381  \\

\textBF{Dev}     & 5,841                               & 209                      & 734                      & 951                      & 1,315                      & 2,070                   & 1,046 & 382  \\

\bottomrule

\end{tabular}
\end{adjustbox}
\caption{Data statistics of KLUE-MRC dataset for long question answering. We count the sum of character lengths of a question and a context.}
\label{tab:kluemrc_statistics}
\end{table}

\begin{table}[t]
\centering
\begin{adjustbox}{width=1\columnwidth}
\begin{tabular}{lccc}

        \toprule

        \multicolumn{4}{c}{\textBF{Question Answering}} \\

        \midrule
        
        \multicolumn{1}{c}{\textBF{Model}} & \textBF{512}                  & \textBF{1024}                 & \textBF{2048}                 \\
        
        \midrule
        
        KLUE-RoBERTa-base         & 67.80/73.73          & -                    & -           \\
        KLUE-RoBERTa-large        & \underline{75.35}/\underline{80.97}          & -        & -       \\

        \midrule

        KoBigBird-base            & 67.32/73.37          & 69.13/75.36          & 68.86/74.20         \\
        KoBigBird-large (ours)    & \textBF{75.57}/\textBF{81.44} & \textBF{73.05}/\textBF{79.06} & \textBF{73.36}/\textBF{79.38}    \\
        $-$ TAPER                    & - & \underline{70.16}/\underline{76.11}          & \underline{71.61}/\underline{77.59} \\
        
        \bottomrule

\end{tabular}
\end{adjustbox}
\caption{The EM and ROUGE-W scores on extractive question answering with different input lengths. Our models are able to handle the input sequence longer than 512. KoBigBird-base remains stable on longer text inputs, and KoBigBird-large scores the best in all input lengths. Without the TAPER method, the model performance decreases, showing its efficacy.}
\label{tab:long_qa}
\end{table}

\subsection{Question Answering for Longer Context}

We venture to address the machine reading comprehension (MRC) task for a longer text-pair NLU task, which derives an appropriate response through the extraction of pertinent context spans answering a question.

Despite the availability of Korean datasets such as KorQuAD 1.0 \cite{lim2019korquad1} and KorQuAD 2.0 \cite{kim2019korquad}, modeled after the notable English SQuAD dataset \cite{rajpurkar2016squad}, they presented significant limitations. The former offers too short input lengths while the latter employs HTML formats, so both are unfit for evaluating long input NLU. To circumvent these drawbacks, we select the KLUE-MRC dataset for our experiment, discussed in Section \ref{subsec:klue_benchmark}, characterized by its adequate input length primarily composed of natural language. We provide the data statistics of KLUE-MRC in Table \ref{tab:kluemrc_statistics}.

The experimental results for MRC are shown in Table \ref{tab:long_qa}. We evaluated the performance with Exact Match (EM) and ROUGE-W scores. Similar to the single document classification task, KoBigBird-large exhibits the best performance across all length segments. It demonstrates superior performance compared to KLUE-RoBERTa-large, illustrating the efficacy of adjustments made in the full attention mode.

While KoBigBird-base shows improved performance with inputs longer than 512 tokens, KoBigBird-large in spite of TAPER slightly regresses because it is crucial to find the exact positions of pertinent spans in the MRC task. Nonetheless, it still outperforms other models.

Our ablation study, without TAPER applied, shows a performance drop within 3\% points.
This indicates that employing TAPER with KoBigBird-large is effective, particularly when handling long inputs, thereby affirming its instrumental role in enhancing the performance in processing extensive texts.

\section{Conclusion}
\label{sec:conclusion}
NLU modules remain highly applicable in fields where neural network throughput and output regularity are critical \cite{yamada2021efficient, baradaran2022survey}. By presenting KoBigBird-large, enabling the simultaneous achievement of state-of-the-art performance and long input processing, this paper contributes to the Korean research community.

\section*{Ethics Statement}

In this study, we ensure strict adherence to ethical considerations, particularly in environmental sustainability and data privacy, following the best practices in AI research.

We minimize environmental impact by using transformation-only on publicly available models, negating further pretraining on parameters and reducing computational load.

For data privacy, we apply pseudonymization techniques during preprocessing, anonymizing all identifiable information in our training corpus. We responsibly transfer parameters from models trained on this pseudonymized corpus, aligning with our dedication to anonymity and ethical AI usage.

Our research methodology, upholding transparency, accountability, and privacy, represents our commitment to the highest ethical conduct and we welcome constructive discourse for continuous improvement.

\section*{Limitations}

The position embeddings of KoBigBird-large, expanded using the TAPER technique, exhibit lower extrapolation performance compared to embeddings obtained through more resource-intensive pretraining. We have endeavored to minimize unnecessary training by transforming the parameters of the existing model, aiming to reduce the carbon footprint associated with the pretraining process. However, this approach has led to a trade-off in the representativeness of the position embeddings.

\section*{Risks}

If users become overly dependent on the model's predictions or suggestions, they might not critically consider or evaluate the generated content, which could lead to the propagation of misinformation or skewed perspectives. There is a potential security risk of adversarial attacks where malicious actors could manipulate the model's output for their purposes. 

\section*{Licenses}
KoBigBird-large is distributed under the terms of the CC BY-SA 4.0 license in accordance with KLUE's licensing policy. This grants open-source users the freedom to copy, redistribute, alter, and build upon the material for any purpose, including commercial endeavors, provided that they distribute their derivative works under an identical license (CC BY-SA 4.0). We anticipate that this approach will significantly enhance future endeavors in NLP research and development.


\bibliography{anthology,custom}
\bibliographystyle{acl_natbib}


\end{document}